\documentclass[twoside,11pt]{article}
\usepackage{jair, theapa, rawfonts}

\usepackage{graphicx}
\usepackage{subcaption}
\usepackage{amsmath}
\usepackage{amsthm}
\usepackage{booktabs}
\usepackage{algorithm}
\usepackage{algorithmic}
\usepackage{bbm}
\usepackage{amssymb}
\usepackage{nicefrac}      
\usepackage{microtype}     
\usepackage{balance}
\usepackage[T1]{fontenc}
\usepackage{color}

\DeclareMathOperator*{\argmax}{argmax}

\newtheorem{theorem}{Theorem}
\newtheorem{definition}{Definition}
\newtheorem{lemma}{Lemma}

\newtheorem{remark}{Remark}

\ShortHeadings{MSV Policy Optimization via Risk-Averse RL}
{Ma, Ma, Xia, \& Zhao}
\firstpageno{1}

\begin{document}

\title{Mean-Semivariance Policy Optimization via Risk-Averse Reinforcement Learning}

\author{\name Xiaoteng Ma \email ma-xt17@mails.tsinghua.edu.cn \\
        \addr Department of Automation, Tsinghua University,\\
        Beijing, 100086, P. R. China
        \AND
       \name Shuai Ma \email mash35@mail.sysu.edu.cn \\
       \addr School of Business, Sun Yat-sen University,\\ Guangzhou, 510275, P. R. China
       \AND
       \name Li Xia \email xiali5@sysu.edu.cn \\
       \textit{(Corresponding author)} \\
       \addr School of Business, Sun Yat-sen University,\\ Guangzhou, 510275, P. R. China
       \AND
       \name Qianchuan Zhao \email zhaoqc@tsinghua.edu.cn \\
        \addr Department of Automation, Tsinghua University,\\
        Beijing, 100086, P. R. China}


\maketitle

\begin{abstract}
Keeping risk under control is often more crucial than maximizing expected rewards in real-world decision-making situations, such as finance, robotics, autonomous driving, etc. The most natural choice of risk measures is variance, which penalizes the upside volatility as much as the downside part. Instead, the (downside) semivariance, which captures the negative deviation of a random variable under its mean, is more suitable for risk-averse proposes. This paper aims at optimizing the mean-semivariance (MSV) criterion in reinforcement learning w.r.t. steady reward distribution. Since semivariance is time-inconsistent and does not satisfy the standard Bellman equation, the traditional dynamic programming methods are inapplicable to MSV problems directly. To tackle this challenge, we resort to Perturbation Analysis (PA) theory and establish the performance difference formula for MSV. We reveal that the MSV problem can be solved by iteratively solving a sequence of RL problems with a policy-dependent reward function. Further, we propose two on-policy algorithms based on the policy gradient theory and the trust region method. Finally, we conduct diverse experiments from simple bandit problems to continuous control tasks in MuJoCo, which demonstrate the effectiveness of our proposed methods.
\end{abstract}

\section{Introduction}

Reinforcement learning (RL) has shown great promise in solving complex decision problems, such as Go~\shortcite{silver2017mastering}, video games~\shortcite{berner2019dota,vinyals2019grandmaster} and dexterous robotic control~\shortcite{nagabandi2020deep}. Learning by trial and error, RL enables an agent to maximize its accumulated expected rewards through interaction with a simulator. However, RL deployment in real-world scenarios is still challenging and unreliable~\shortcite{garcia2015comprehensive,dulac2019challenges}. One of the reasons is that real decision-makers need to consider multi-objective functions. The desired policy should perform well for broader metrics, not just for expectation. That raises the demand of \emph{risk-sensitive learning}, which aims at balancing the return and risk in face of uncertainty. 

The risk-sensitive decision-making has been widely studied beyond the scope of RL, which can be traced back to the \emph{mean-variance} (MV) optimization theory established by~\shortciteA{markowitz1952portfolio}. Variance, which captures the fluctuation and concentration of random variables, is a natural choice of the risk measure. As Markowitz only considers the single-period problem, many studies focus on extending the results to multi-period scenarios, from stochastic control~\shortcite{li2000optimal} to Markov decision process~\shortcite{sobel1982variance,filar1989variance}. However, the variance of a multi-period problem depends on the average value of the whole process. It breaks the essential property of dynamic programming---time consistency and makes it hard to design model-free learning algorithms under the standard RL framework. Developing an efficient algorithm to optimize MV is still an ongoing topic in the RL community~\shortcite{xie2018block,bisi2019risk,xia2020risk,zhang2021mean,ma2022unified,MA2022100165}. 

\begin{figure}[t]
    \centering
    \includegraphics[width=0.5\linewidth]{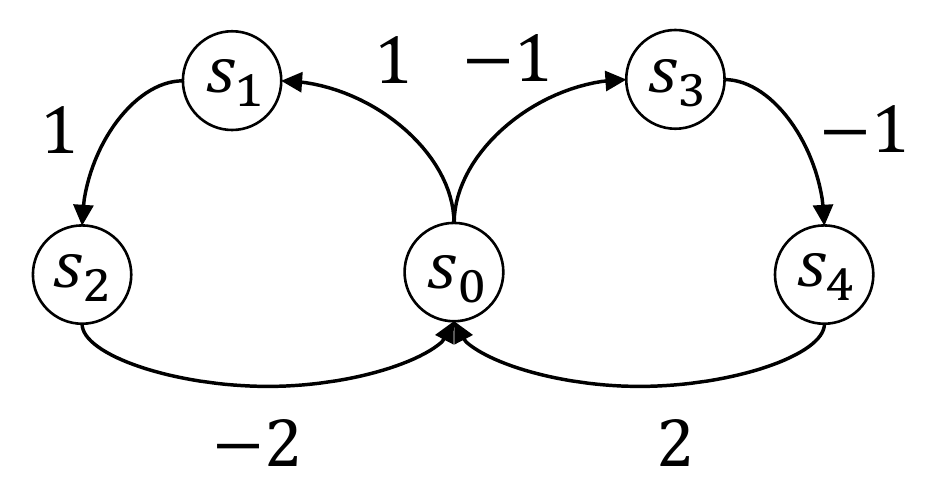}
    \caption{A toy example illustrates the effect of MSV. We refer the policy going left as $l$ and the other as $r$. Two policies have the same average return $\eta^l=\eta^r=0$ and the same variance $\zeta^l=\zeta^r=2$. However, since the semivariance $\zeta_-^l = 4/3 > \zeta_-^r = 2 /3$, the policy going right has a smaller (downside) semivariance. It shows that MSV enables to avoid extreme costs compared with MV.}
    \label{fig:schematic}
\end{figure}

While MV analysis is the most widely applied risk-return analysis in practice, variance metric is questionable as a risk measure. As a measure of volatility, variance penalizes upside deviations from the mean as much as downside deviations. It could be problematic as the upside deviation comes from the higher return which is desirable. In general, the outcome distributions in the real world are often asymmetrical, such as the ones in the stock market~\shortcite{estrada2007mean,bollerslev2020good}, suggesting that we should control the ``good'' and ``bad'' volatility separately. Hence,~\shortciteA{markowitz1959portfolio} presents the \emph{mean-semivariance} (MSV) as an alternative measure, which only penalizes the ``bad'' volatility, performing as a downside risk indicator. Even if the distribution is symmetrical, optimizing MSV is at least effective as optimizing MV. To better illustrate the difference between variance and semivariance, we construct a simple MDP example shown in Figure~\ref{fig:schematic}. The two policies result in two reward distributions symmetrically, for which variances are indistinguishable. However, the policy going right is preferred since it results in a lower semivariance.

Though MSV is a more plausible measure of risk, optimizing MSV is even more complicated than MV. It inherits time inconsistency from variance and introduces a truncation function of mean, making the analysis non-trivial. Due to the complexity of this objective, existing works consider a subset of problems restricted with a fixed mean~\shortcite{wei2019mean} or heuristic algorithms for MSV~\shortcite{yan2007multi,zhang2012possibilistic,liu2015multi,chen2019multi}. To the best of our knowledge, there are currently no relevant studies on MSV in the RL literature.

In this paper, we aim to fill the gap of the previous study on the single-period MSV problem and extend the static methods to online RL algorithms. To achieve that, we resort to Perturbation Analysis (PA) theory~\cite{cxr} (also called the sensitivity-based optimization theory or the relative optimization theory) for Markov systems, which lays the basis of many efficient RL methods, such as TRPO~\shortcite{trpo}, CPO~\shortcite{cpo} and MBPO~\shortcite{janner2019trust}. The contributions of our work are threefold. \emph{Firstly}, instead of constructing a Bellman operator, we establish the MSV performance difference formula of two policies (see Section~\ref{sec:PA} for details). The result indicates that the performance difference can be decomposed into two parts: the improvement corresponding to a reward function depending on the current policy and the average performance change from the current to the updated one. \emph{Second}, we iteratively optimize MSV by considering the shift in mean locally and constructing a surrogate reward function. The framework is shown in Algorithm~\ref{alg:MSV}. Under this framework, we develop two algorithms based on the policy gradient theory and the trust region method, respectively. We show that optimizing the surrogate reward function in the trust region has a similar performance lower bound with the standard TRPO, which guarantees monotonic improvement if the trust region is tight. \emph{Finally}, we conduct diverse experiments to examine the effectiveness of our proposed methods, including a bandit problem, a tabular portfolio management problem, and robotic control tasks based on MuJoCo. The results demonstrate that the proposed algorithms successfully improve the performance under the criterion of MSV, which is better than standard RL from a risk-averse perspective.


 

\section{Related Work}
 Below we briefly review the literature about optimization of MSV and other risk measures.

\begin{algorithm}[t]
\caption{The framework of MSV optimization}  
\label{alg:MSV}  
\begin{algorithmic}

\STATE Initialize policy as $\mu$
\REPEAT
\STATE Evaluate $\mu$ and get $\eta$ (cf. Equation~\ref{equ:eta}) and $\eta_-$ (cf. Equation~\ref{equ:semimean})
\STATE Set reward function as $g = (1 + 2 \beta \eta_-) r - \beta (r - \eta)_-^2$
\STATE $\mu \gets {\rm POLICY\_UPDATE}(\mu, g)$
\UNTIL {$\mu$ converges}
\end{algorithmic}  
\end{algorithm}

\subsection{Mean-Semivariance}
MSV is first introduced by~\shortciteA{markowitz1959portfolio} as an alternative to MV. Thereafter, many researchers study portfolio selection problems by employing the semivariance as the risk measure~\shortcite{markowitz1993computation,hogan1974toward,choobineh1986simple,briec2009multi}, most of which are limited to the single-period problem.
Due to the complexity of MSV, previous studies on MSV in multi-period problems resort to heuristic methods, such as fuzzy systems and genetic algorithms~\shortcite{yan2007multi,zhang2012possibilistic,liu2015multi,chen2019multi}.
\shortciteA{wei2019mean} studies a special case of MSV in the continuous-time MDP, where the mean of the discounted total cost is equal to a given function. Another stream of research~\shortcite {tamar2016sequential,shapiro2021lectures} studies semideviation instead of semivariance. As standard deviation is an alternative to variance, semideviation is considered an alternative to semivariance. The main benefit of mean-semideviation (MSD) is that it satisfies the property ``coherent,'' and hence it can be written in a Bellman form~\shortcite{ruszczynski2010risk}. However, the additional square operation makes optimizing MSD with a data-driven approach non-trivial. We leave the optimization of MSD in RL as future work. Furthermore, maximizing the upside semivariance could improve the exploration ability~\shortcite{mavrin2019distributional,ma2020dsac,zhou2020non}, showing the potential of MSV from an opposite perspective.

\subsection{Mean-Variance}
Since MSV is highly related to MV, in this part, we summarize the works on MV in Markov decision processes (MDPs) and RL. Based on the definition of variance in the framework of MDPs, the existing studies on variance can be broadly divided into two categories. One stream of works~\shortcite{sobel1982variance,tamar2012policy,prashanth2016variance,xie2018block} concern the variance of total return $R = \sum_{t=0}^\infty \gamma^t r_t$ under the initial state distribution, i.e., $\mathbb{V}_{\pi_0}(R)$ where $\gamma$ is the discount factor, $\pi_0$ is the initial state distribution and $r_t$ is the reward at the stage $t$. This definition concerns the risk of total rewards at the final stage, while we are more concerned about long-term volatility in practical problems. Hence, the long-run variance~\shortcite{filar1989variance,chung1994mean,gosavi2014variance,xia2016optimization,xia2020risk,bisi2019risk,zhang2021mean}, also known as the steady-state variance, is proposed to describe the variance of the steady reward distribution. 
The long-run variance is defined by $\lim_{T \to \infty} \frac{1}{T} \mathbb{E}_{\pi_0, \mu} \left[ \sum_{t=0}^{T-1} \left(r(s_t, a_t) -\eta^\mu \right)^2 \right]$ (cf. Equation~\ref{equ:def_zeta}), where $\eta^\mu$ is the long-run average of policy $\mu$. Since the average reward $\eta^\mu$ depends on the current policy, it breaks the time-consistency. To handle this problem, ~\shortciteA{xia2016optimization,xia2020risk} derives a variance performance difference formula with PA and proposes a policy iteration algorithm that is guaranteed to converge to a local optimum. In this paper, we adopt a similar definition of~\citeauthor{xia2016optimization,xia2020risk}'s work and extend the formulation from MV to MSV.   

\subsection{Other Risk Measures}
Besides the MV and MSV, other risk measures capture different features of the return distribution. A classical risk measure in optimal control is exponential utility~\shortcite{howard1972risk,borkar2002risk,fei2020risk}. The exponential utility enjoys a product form of the Bellman equation. Therefore the corresponding value-based algorithms such as Q-learning are well-developed. While the exponential Bellman equation is elegant in theory, it poses some computational problems as the exponential values are often too large to be numerically calculated. Another famous risk measure is Conditional Value at Risk (CVaR), defined as the average value under the $\alpha$-quantile. Many existing methods~\shortcite{nemirovski2007convex,chow2014algorithms,tamar2015optimizing,chow2015risk,chow2017risk} optimize CVaR as the objective or constraints. The main difference between CVaR and MSV is that CVaR puts even weights for the events under a certain threshold, while the importance of the extreme values on the concerned side increases quadratically in MSV. We refer to~\shortciteauthor{delage2019dice}'s work~\citeyear{delage2019dice} for more discussion on the connection of different risk measures.

\section{Preliminaries}
In this paper, we focus on the infinite-horizon discrete-time MDP as $\mathcal{M} = \langle \mathcal{S}, \mathcal{A}, r, P, \pi_0 \rangle$, where $\mathcal{S}$ denotes the state space, $\mathcal{A}$ denotes the action space, $r: \mathcal{S} \times \mathcal{A} \mapsto [-R_{\rm max}, R_{\rm max}]$ denotes a bounded reward function and $P: \mathcal{S} \times \mathcal{A} \mapsto \Delta(\mathcal{S})$ is the transition matrix and $\pi_0 \in  \Delta(\mathcal{S})$ denotes the initial state distribution. We assume that all the involved MDPs are ergodic. Let $\mu: \mathcal{S} \mapsto \Delta(\mathcal{A}) $ denote a Markovian randomized policy and $\Pi$ denote the randomized policy space.

We are interested in the long-run average reward
\begin{equation}\label{equ:eta}
    \eta^\mu := \lim_{T \to \infty} \frac{1}{T} \mathbb{E}_{\pi_0, \mu} \left[ \sum_{t=0}^{T-1} r(s_t, a_t) \right],
\end{equation}
where $\mathbb{E}_{\pi_0, \mu}$ stands for the expectation with $s_0 \sim \pi_0, a_t \sim \mu( \cdot \mid s_t), s_{t+1} \sim P( \cdot \mid s_t, a_t)$. Note that $\eta^\mu$ is independent of $\pi_0$ when $T \to \infty$. With $\pi$ denoting the steady-state distribution, it is convenient to rephrase the long-run average reward as
\begin{equation}
    \eta^\mu := \mathbb{E}_{s \sim \pi, a \sim \mu} \left[ r(s, a) \right].
\end{equation}
The variance and semivariance w.r.t. $\mu$ are defined by
\begin{align} \label{equ:def_zeta}
    \zeta^\mu &:= \lim_{T \to \infty} \frac{1}{T} \mathbb{E}_{\pi_0, \mu} \left[ \sum_{t=0}^{T-1} \left(r(s_t, a_t) -\eta^\mu \right)^2 \right], \\
    \zeta_-^\mu &:= \lim_{T \to \infty} \frac{1}{T} \mathbb{E}_{\pi_0, \mu} \left[ \sum_{t=0}^{T-1} \left(r(s_t, a_t) -\eta^\mu \right)_-^2 \right],
\end{align}
where $(\cdot)_-:=\min\{0, \cdot\}$.
In this paper, we focus on the mean-semivariance criterion, 
\begin{equation*}
    \xi_-^\mu := \eta^\mu - \beta \zeta_-^\mu, 
\end{equation*}
where $\beta \geq 0$ is the parameter for the trade-off between mean and semivariance. Analogously, when mean-variance criterion is mentioned, we mean $\xi^\mu := \eta^\mu - \beta \zeta^\mu$.

We further respectively define the state-value function, action-value function, and advantage function for average reward as 
\begin{align*}
V_{\eta}^\mu(s) &:= \mathbb{E}_{\mu} \left[\sum_{t=0}^{\infty} (r(s_{t}, a_{t}) - \eta^\mu) \mid s_0=s \right], \notag \\
Q_{\eta}^\mu(s, a) &:= \mathbb{E}_{\mu} \left[\sum_{t=0}^{\infty} (r(s_{t}, a_{t}) - \eta^\mu) \mid s_0=s, a_0=a \right], \notag \\
A_{\eta}^\mu(s, a) &:= Q_{\eta}^\mu(s, a) - V_{\eta}^\mu(s).
\end{align*}
Similarly, the value functions for semivariance are defined as
\begin{align*}
V_{\zeta_-}^\mu(s) &:= \mathbb{E}_{\mu} \left[\sum_{t=0}^{\infty} \left( (r(s_{t}, a_{t}) - \eta^\mu )_-^2 - \zeta_-^\mu \right) \mid s_0=s \right], \notag \\
Q_{\zeta_-}^\mu(s, a) &:= \mathbb{E}_{\mu} \left[\sum_{t=0}^{\infty} \left((r(s_{t}, a_{t}) - \eta^\mu )_-^2 - \zeta_-^\mu \right) \mid s_0=s, a_0=a \right], \notag \\
A_{\zeta_-}^\mu(s, a) &:= Q_{\zeta_-}^\mu(s, a) - V_{\zeta_-}^\mu(s).
\end{align*}

For notation simplicity, we will omit the superscript ``$\mu$'' when the context is clear, e.g., the average rewards $\eta^\mu,\eta^{ \mu^\prime}$ are written as $\eta, \eta^\prime$ instead. When $r$ is mentioned, we omit $(s, a)$ and use $r$ in short.

Before our analysis of MSV, we briefly review the average-reward policy gradient theorem and the trust region theorem.
\begin{theorem}[Average-Reward Policy Gradient by~\shortciteR{sutton2018reinforcement}]
For a policy $\mu$ parameterized by $\theta$, we have
    \begin{equation*}
        \nabla_\theta \eta = \mathbb{E}_{s \sim \pi, a \sim \mu} [  \nabla_\theta \log \mu(a \mid s) A_\eta^\mu(s, a) ].
    \end{equation*}
\end{theorem}

\begin{theorem}[Average-Reward Trust Region Policy Optimization by~\shortciteR{pmlr-v139-zhang21q,ijcai2021-385}]
Consider the following problem, 
\begin{align} \label{equ:mean_trpo_problem}
    & \max_{\mu_\theta} \mathcal{L}^{\mu}(\mu_\theta), \\
    & \ \mathrm{s.t. } \ \mathbb{E}_{s \sim \pi} D_{\rm TV}(\mu_\theta( \cdot \mid s) \parallel \mu( \cdot \mid s) ) \leq \epsilon_\mu, \notag
\end{align}
where
\begin{align}
    \mathcal{L}^{\mu}(\mu_\theta) := \mathbb{E}_{s \sim \pi, a \sim \mu_\theta} \left[ A_\eta^{\mu}(s,a) \right].
\end{align}
Denote $\mu^\prime$ as the solution of the above problem. The following bound holds:
\begin{equation} \label{equ:mean_lower_bound} 
    \eta^\prime - \eta \geq \mathcal{L}^{\mu}(\mu^\prime) - 2  (\kappa^\prime -1) \epsilon_\eta \epsilon_\mu, 
\end{equation}
where $\epsilon_\eta = \max_s | \mathbb{E}_{a \sim \mu^\prime} [A_\eta^\mu(s,a)] |$ and $\kappa^\prime$ is Kemeny's constant under $\mu^\prime$.
\end{theorem}

\section{Perturbation Analysis}
\label{sec:PA}
In this section, we derive the \emph{MSV performance difference formula} (MSVPDF), where the core concept---performance difference formula---comes from the PA for Markov systems, also called the sensitivity-based optimization theory. With the aid of MSVPDF, we obtain the necessary optimality condition for the MSV problem. It also lays the basis for developing optimization algorithms (see Section~\ref{sec:algo}), such as the policy gradient method and the trust region method. For readers unfamiliar with PA, we provide a brief review of the theory in Appendix~\ref{app:PA}.

\subsection{Performance Difference Formula}
MSVPDF is formally stated below.
\begin{theorem}
    For any two policies $\mu, \mu^\prime \in \Pi$, we have
    \begin{equation}
        \label{equ:pdf_sample}
        \xi^{\prime}_- - \xi_- = \mathbb{E}_{s \sim \pi^\prime, a \sim \mu^\prime} [A_{\eta}^\mu(s, a) - \beta A_{\zeta_-}^\mu(s, a)] - \beta \mathbb{E}_{s \sim \pi^\prime, a \sim \mu^\prime} [(r - \eta^{\prime})^2_- - (r - \eta)^2_- ].
    \end{equation} 
\end{theorem}
\begin{proof}
To decompose the policy performance with the policy-dependent reward, we first introduce a pseudo mean $\lambda$. We analyze the policy difference with the pseudo mean and corresponding pseudo reward function and then turn it into the true mean by letting $\lambda = \eta$.

With a pseudo mean $\lambda$, we transform the original problem into a standard MDP with reward function
\begin{equation} \label{equ:f}
    f(s, a) := r - \beta(r - \lambda)_-^2.
\end{equation}

We obtain a pseudo mean-semivariance objective by optimizing this pseudo reward-function,
\begin{equation*}
    \xi_{\lambda, -} := \xi_{\lambda, -}^\mu = \mathbb{E}_{s \sim \pi, a \sim \mu} \left[ f(s, a)  \right].
\end{equation*}
By definition, we have
\begin{align*}
    & \xi_{-} - \xi_{\lambda, -} = \mathbb{E}_{s \sim \pi, a \sim \mu} \left[ r - \beta(r-\eta)_-^2 - f(s, a) \right].
\end{align*}
Since the pseudo reward is independent of the policy, we can write its performance difference formula directly~\shortcite[Chapter 2]{cxr}:
\begin{equation} \label{equ:diff_pesudo}
    \xi_{\lambda, -}^\prime - \xi_{\lambda, -} = \mathbb{E}_{s \sim \pi^\prime, a \sim \mu^\prime} [A_f^\mu(s, a)],
\end{equation}
where $A_f^\mu(s, a)$ is the pseudo advantage with $f$ as the reward function. With the aid of Equation~\ref{equ:diff_pesudo}, we can derive the performance difference formula of $\xi_-$ as
\begin{align*}
    \xi_-^\prime - \xi_- &=  (\xi_{\lambda, -}^\prime - \xi_{\lambda, -})  + (\xi_-^\prime - \xi_{\lambda, -}^\prime) + (\xi_{\lambda, -} - \xi_-) \\
    &=\mathbb{E}_{s \sim \pi^\prime, a \sim \mu^\prime} [A_f^\mu(s, a)] - \beta \mathbb{E}_{s \sim \pi^\prime, a \sim \mu^\prime} \left[ (r - \eta^\prime)_-^2 - (r - \lambda)_-^2  \right] \\
    & \quad - \beta \mathbb{E}_{s \sim \pi, a \sim \mu} \left[ (r - \lambda)_-^2 - (r - \eta)_-^2  \right]. 
\end{align*}
Finally, by setting $\lambda = \eta$, we arrive at
\begin{equation} \label{equ:msv_diff}
    \xi_-^\prime - \xi_- = \mathbb{E}_{s \sim \pi^\prime, a \sim \mu^\prime} [A_f^\mu(s, a)] - \beta \mathbb{E}_{s \sim \pi^\prime, a \sim \mu^\prime} \left[ (r - \eta^\prime)_-^2 - (r - \eta)_-^2  \right], \notag
\end{equation}
which is the same as Equation~\ref{equ:pdf_sample} if we explicitly calculate the advantage function with reward function $f$ and $\lambda=\eta$.
\end{proof}
The MSVPDF in Equation~\ref{equ:pdf_sample} or Equation~\ref{equ:msv_diff} claims that the MSV improvement can be divided into two parts. The first term in Equation~\ref{equ:msv_diff} is a standard MDP with $f$ as the reward function, and the second term is caused by the perturbation of the mean. It clearly quantifies the difficulty of solving the MSV problem, i.e., \emph{the policy-dependent reward function breaks down the time-consistent nature of MDPs}. Meanwhile, it also shows us the standard MDP algorithm such as policy iteration (PI) is unavailable. A PI-like algorithm may be efficient in improving the first term, but the sign of the remaining term (dependent on $\eta^\prime$) is unpredictable. It suggests that we need novel tools to guarantee policy improvement.  

\subsection{Performance Derivative Formula}
\label{sec:PDF2}
While Equation~\ref{equ:msv_diff} describes the performance difference between any two policies, we still need the local structure of the MSV problem to guide the direction of optimization. Following the line of the last part, we present the \emph{MSV performance derivative formula} in this subsection, which describes the performance derivative at $\mu$ towards another policy $\mu^\prime$. 
\begin{theorem}
Given any two policies $\mu, \mu^\prime \in \Pi$, we consider a mixed policy $\mu^\nu$, 
\begin{equation*}
    \mu^\nu(a \mid s) = (1 - \nu) \mu(a \mid s) + \nu \mu^\prime(a \mid s),
\end{equation*}
where the action follows $\mu$ with probability $1 - \nu$, and follows $\mu^\prime$ with probability $\nu$ for $\nu \in [0, 1]$. We have
\begin{equation*}
    \frac{\mathrm{d} \xi_-}{\mathrm{d} \nu} = \mathbb{E}_{s \sim \pi, a \sim \mu^\prime} [ (1 + 2 \beta \eta_-)A_{\eta}^\mu(s, a) - \beta A_{\zeta_-}^\mu(s, a)].
\end{equation*}
\end{theorem}

\begin{proof}
From MSVPDF, we obtain the difference for $\mu, \mu^\nu$,
\begin{equation*}
     \xi_-^\nu - \xi_- = \mathbb{E}_{s \sim \pi^\nu, a \sim \mu^\nu} [A_f^\mu(s, a)] - \beta \mathbb{E}_{s \sim \pi^\nu, a \sim \mu^\nu} \left[ (r - \eta^\nu)_-^2 - (r - \eta)_-^2  \right],
\end{equation*}
where $\eta^\nu:=\eta^{\mu^\nu}$.
Taking the derivative w.r.t. $\nu$ and letting $\nu \to 0$, we obtain the performance derivative formula. To simplify the derivation, we denote the terms on the right-hand side as
\begin{align*}
    h_1(\nu) &= \mathbb{E}_{s \sim \pi^\nu, a \sim \mu^\nu} [A_f^\mu(s, a)], \\
    h_2(\nu) &= \mathbb{E}_{s \sim \pi^\nu, a \sim \mu^\nu} \left[ (r - \eta^\nu)_-^2 - (r - \eta)_-^2  \right].
\end{align*}
Then $\xi_-^\nu - \xi_- = h_1(\nu) - \beta h_2(\nu)$. Specifically, we have
\begin{align*}
    h_1(\nu) &= \mathbb{E}_{s \sim \pi^\nu} [(1 - \nu) \mathbb{E}_{a \sim \mu} [A_f^\mu(s, a)] + \nu \mathbb{E}_{a \sim \mu^\prime} [A_f^\mu(s, a)]] \\
    &=  \nu \mathbb{E}_{s \sim \pi^\nu, a \sim \mu^\prime} [A_f^\mu(s, a)],
\end{align*}
where the last equality follows that $\mathbb{E}_{a \sim \mu} [A_f^\mu(s, a)] = 0$. Since $\lim_{\nu \to 0} \pi^\nu = \pi$, we obtain
\begin{equation*}
    \frac{\mathrm{d} h_1}{\mathrm{d} \nu} = \mathbb{E}_{s \sim \pi, a \sim \mu^\prime} [A_f^\mu(s, a)].
\end{equation*}
Next, we differentiate $(r - \eta)_-^2$,
\begin{equation} \label{equ:diff_rew}
\begin{aligned}
    \frac{\mathrm{d} (r - \eta)_-^2}{\mathrm{d} \nu} &=  2 (r - \eta)_- \frac{\mathrm{d} (r - \eta^\nu)_- }{\mathrm{d} \nu} \\
    & \overset{(\romannumeral 1)}{=} -2 (r - \eta)_- \mathbbm{1}(r < \eta )\frac{\mathrm{d} \eta}{\mathrm{d} \nu} \\
    & \overset{(\romannumeral 2)}{=} -2 (r - \eta)_- \frac{\mathrm{d} \eta}{\mathrm{d} \nu},
\end{aligned}    
\end{equation}
where $(\romannumeral 1)$ follows $\frac{{\mathrm d (x)_-} }{{\mathrm d x}} = \mathbbm{1}(x < 0)$, and $(\romannumeral 2)$ comes from $(r - \eta)_- \mathbbm{1}(r < \eta) = (r - \eta)_-$. Thus, we have
\begin{align*}
    \frac{\mathrm{d} h_2}{\mathrm{d} \nu}&= \lim_{\nu \to 0} \frac{1}{\nu} \sum_s \pi^\nu(s) \sum_a \mu^\nu(a \mid s) \left[ (r - \eta^\nu)_-^2 - (r - \eta)_-^2  \right] \\
    &= \lim_{\nu \to 0} \sum_s \pi^\nu(s) \sum_a \mu^\nu(a \mid s) \frac{(r - \eta^\nu)_-^2 - (r - \eta)_-^2}{\nu} \\
    &= \sum_s \pi(s) \sum_a \mu(a \mid s) \frac{\mathrm{d} (r - \eta)_-^2}{\mathrm{d} \nu} \\
    &= \sum_s \pi(s) \sum_a \mu(a \mid s) \left[-2 (r - \eta)_- \frac{\mathrm{d} \eta}{\mathrm{d} \nu} \right] \\
    &= -2 \eta_- \frac{\mathrm{d} \eta}{\mathrm{d} \nu}. 
\end{align*}
Here we define the \emph{semimean} $\eta_-$ as
\begin{equation} \label{equ:semimean}
    \eta_- := \eta_-^\mu = \mathbb{E}_{s \sim \pi, a \sim \mu} [(r - \eta)_-],
\end{equation}
which is the downside expectation of rewards under $\pi$.
From the standard result of PA~\shortcite[Chapter 2]{cxr}, we have
\begin{equation*}
    \frac{\mathrm{d} \eta}{\mathrm{d} \nu} = \mathbb{E}_{s \sim \pi, a \sim \mu^\prime} [A_{\eta}^\mu(s, a)].
\end{equation*}
Putting the above relationships together, we obtain
\begin{align*}
    \frac{\mathrm{d} \xi_-}{\mathrm{d} \nu} &=
    \frac{\mathrm{d} h_1}{\mathrm{d} \nu} - \beta \frac{\mathrm{d} h_2}{\mathrm{d} \nu} \\
    &=\mathbb{E}_{s \sim \pi, a \sim \mu^\prime} [ A_f^\mu(s, a)] + 2 \beta \eta_-
    \frac{\mathrm{d} \eta}{\mathrm{d} \nu} \\
    &= \mathbb{E}_{s \sim \pi, a \sim \mu^\prime} [ A_f^\mu(s, a) + 2 \beta \eta_- A_\eta^\mu(s, a)] \\
    &= \mathbb{E}_{s \sim \pi, a \sim \mu^\prime} [ (1 + 2 \beta \eta_-)A_{\eta}^\mu(s, a) - \beta A_{\zeta_-}^\mu(s, a)].    
\end{align*}
\end{proof}

The above equality indicates that the performance derivative is related to another reward function w.r.t. $f$ (cf. Equation~\ref{equ:f}): 
\begin{align} \label{equ:rew_g}
    g(s, a) &:= f(s, a) + 2 \beta \eta_- r \\
    &=(1 + 2 \beta \eta_-) r - \beta (r - \eta)_-^2,
\end{align}
and the derivative formula can be written as
\begin{equation} \label{equ:derivative formula}
    \frac{\mathrm{d} \xi_-}{\mathrm{d} \nu} = \mathbb{E}_{s \sim \pi, a \sim \mu^\prime} [ A_g^\mu(s, a)],
\end{equation}
where $A_g^\mu(s, a)$ is the advantage function w.r.t. $g$.

With the performance derivative formula, we define the local optimum for MSV and present the necessary condition for MSV optimality.

\begin{definition}
For a policy $\mu$, $\exists \bar{\nu} \in (0, 1)$ and we always have $\xi_-^\mu \geq \xi_-^\nu, \forall \nu \in (0, \bar{\nu})$, then we say $\mu$ is a local optimum in the mixed policy space.
\end{definition}

\begin{theorem}
The optimal policy of MSV can be found in the deterministic policy space and satisfies the necessary condition 
\begin{equation*}
    \mu^*(a \mid s) = \delta \left(a \in \argmax_{b \in \mathcal{A}} A^*_g(s, b) \right),
\end{equation*}
which implies that $A_g^*(s,a) \leq 0, \forall s \in \mathcal{S},a \in \mathcal{A}$. Here $\delta$ denotes the Dirac delta function.

\end{theorem}
\begin{proof}
The theorem is a direct result of the derivative formula. The (local) optimality implies that if $\mu$ is a local optimum, we always have $\frac{\mathrm{d} \xi_-}{\mathrm{d} \nu} \leq 0$ for any direction in the policy space. Assuming there is a contradiction, where for a state $s$ there exists $\mu(a \mid s) = \delta(a = a^\prime)$ for any  $ a^\prime \notin \argmax_b A_g^\mu(s, b)$, we can always find a better policy in the mixed policy space along the derivative direction.
\end{proof}

\section{Optimization and Algorithms}
\label{sec:algo}

In this section, we propose two approaches to optimize MSV with the parameterized policy. We firstly extend the policy gradient method to MSV with the pseudo reward function (cf. Equation~\ref{equ:rew_g}) in Section~\ref{sec:PA}. Following the same idea, we propose a trust region method to solve the MSV problem and prove the lower bound for its performance improvement. The two approaches together establish an iterative framework to solve the MSV problem.

\subsection{MSV Policy Gradient Method}

Policy gradient theorem is an essential foundation of modern deep RL algorithms, such as Actor-Critic methods. Here we consider the policy $\mu$ parameterized by $\theta \in \Theta$, which can be implemented with any differentiable function. We first give the MSV Policy Gradient (MSVPG) theory formally as follows.
\begin{theorem} For a policy $\mu$ parameterized by $\theta$, we have
    \begin{equation} \label{equ:msvpg}
        \nabla_\theta \xi_- = \mathbb{E}_{s \sim \pi, a \sim \mu} [  \nabla_\theta \log \mu(a \mid s) A_g^\mu(s, a) ].
    \end{equation}
\end{theorem}
The policy gradient for MSV can be easily proved by PA, which follows the same lines as the derivative formula. For the readers from the DRL community, we also provide an alternative proof based on~\shortcite{sutton2018reinforcement} in the appendix.
\begin{proof}
Consider two policies $\mu, \mu^\prime$ parameterized by $\theta, \theta^\prime$ respectively. Their performance difference is given as
\begin{equation*}
    \xi_-^\prime - \xi_- =  \mathbb{E}_{s \sim \pi^\prime, a \sim \mu^\prime} [A_{f}^\mu(s, a)]
    - \beta \mathbb{E}_{s \sim \pi^\prime, a \sim \mu^\prime} \left[ (r - \eta^\prime)_-^2 - (r - \eta)_-^2  \right]. 
\end{equation*}
Let denote $\Delta \theta = \theta^\prime - \theta$. Similar to the derivation in Section~\ref{sec:PDF2}, we denote the terms of above equation
\begin{align*}
    h_1(\Delta \theta) &= \mathbb{E}_{s \sim \pi^\prime, a \sim \mu^\prime} [A_f^\mu(s, a)], \\
    h_2(\Delta \theta) &= \mathbb{E}_{s \sim \pi^\prime, a \sim \mu^\prime} \left[ (r - \eta^\prime)_-^2 - (r - \eta)_-^2  \right].
\end{align*}
We take the limit of $\xi^{\prime}_- - \xi_-$ by letting $\theta^\prime \to \theta$.
\begin{align*}
    \nabla_\theta h_1
    &= \lim_{\Delta \theta \to 0} \frac{1}{\Delta \theta} \sum_s  \pi^\prime(s) \sum_a \left[ \mu^\prime(a \mid s)  A_f^\mu(s, a) \right] \\
    &\overset{(\romannumeral 1)}{=}  \lim_{\Delta \theta \to 0} \sum_s  \pi^\prime(s) \sum_a \frac{ \mu^\prime(a \mid s) - \mu(a \mid s)}{\Delta \theta} A_f^\mu(s, a) \\
    &= \sum_s \pi(s) \sum_a  \nabla_\theta \mu(a \mid s) A_f^\mu(s, a) \\
    &\overset{(\romannumeral 2)}{=} \mathbb{E}_{s \sim \pi, a \sim \mu} \left[ \nabla_\theta \log \mu(a \mid s) A_f^\mu(s, a) \right],
\end{align*}
where $(\romannumeral 1)$ follows $\mathbb{E}_{a \sim \mu} [A_f^\mu(s, a)]=0$ and $(\romannumeral 2)$ comes from $\nabla_\theta \log \mu(a \mid s) = \dfrac{\nabla_\theta \mu(a \mid s)}{\mu(a \mid s)}$.

Similar to the derivation in Equation~\ref{equ:diff_rew}, we have
\begin{equation*} 
\begin{aligned}
    \nabla_\theta h_2 &= \lim_{ \Delta \theta \to 0} \sum_s \pi^\prime(s)\sum_a \mu^\prime(a \mid s) \frac{ (r - \eta^\prime)_-^2 - (r - \eta)_-^2}{\Delta \theta} \\
    &= \sum_s \pi(s)\sum_a \mu(a \mid s) \lim_{\Delta \theta \to 0} \frac{ (r - \eta^\prime)_-^2 - (r - \eta)_-^2}{\Delta \theta} \\
    &= \sum_s \pi(s)\sum_a \mu(a \mid s) \nabla_\theta (r - \eta)_-^2 \\
    &= \sum_s \pi(s)\sum_a \mu(a \mid s) [-2 (r - \eta)_- \nabla_\theta \eta] \\
    &= -2 \eta_- \nabla_\theta \eta.
\end{aligned} 
\end{equation*}

Since $\nabla_\theta \eta = \mathbb{E}_{s \sim \pi, a \sim \mu} \left[ \nabla_\theta \log \mu(a \mid s) A_\eta^\mu(s, a) \right]$, we combine the results together and give the gradient of $\xi_-$
\begin{align*}
    \nabla_\theta \xi_- &= \nabla_\theta h_1 - \beta \nabla_\theta h_2 \\
    &= \mathbb{E}_{s \sim \pi, a \sim \mu} \left[ \nabla_\theta \log \mu(a \mid s) A_f^\mu(s, a) \right] + 2 \beta \eta_- \nabla_\theta \eta  \\
    &= \mathbb{E}_{s \sim \pi, a \sim \mu} \left[ \nabla_\theta \log \mu(a \mid s) A_f^\mu(s, a) + 2  \beta \eta_- A_\eta^\mu(s, a) \right] \\
    &= \mathbb{E}_{s \sim \pi, a \sim \mu} \left[ \nabla_\theta \log \mu(a \mid s) A_g^\mu(s, a)\right].
\end{align*}
\end{proof}
Here we present an Actor-Critic algorithm based on MSVPG, which is named MSVAC (see Algorithm~\ref{alg:MSVPG}). In addition to the parameterized policy, we maintain another parameterized function $V_\phi$ as the value function. Then, the advantage function is estimated with the generalized advantage estimation (GAE)~\shortcite{gae}. Typically, we have
\begin{equation} \label{equ:gae}
    \hat{A}_g(s_n,a_n) = \sum_{t=n}^{N-1} \lambda^{t-n} \left(g(s_t, a_t) - \hat g + V_\phi(s_{t}) - V_\phi(s_{t+1}) \right),
\end{equation}
where $\lambda$ is the hyper-parameter to trade-off bias and variance, and $\hat g = (1 + 2\beta \hat{\eta}_-)\hat{\eta} -\beta \hat{\zeta}_-$ is the estimation of average surrogate reward function. With $\hat{V}_n = V_\phi(s_n) + \hat{A}_g(s_n, a_n)$ as the target value, we update the value function with
\begin{equation} \label{equ:val_loss}
    \mathcal{L}_V(\phi) := \frac{1}{2N} \sum_{n=0}^{N-1} (V_\phi(s_n) - \hat{V}_n)^2.
\end{equation}

\begin{algorithm}[tb] 
\caption{MSVAC}  
\label{alg:MSVPG}  
    \textbf{Input}: $\alpha, \beta, K, N$
    \begin{algorithmic}[1]
        \STATE Initialize the policy with $\theta$ and the value with $\phi$ randomly.
        \STATE Set $\hat{\eta} = 0$, $\hat{\eta}_- = 0$, $\hat{\zeta}_- = 0$.
        \FOR {$k=1,2,\cdots, K$} 
        \STATE Execute policy $\mu_{\theta}$ for $N$ times to collect $\{(s_n, a_n, r_n, s_{n+1})\}_{n=0}^{N-1}$.
        \STATE Update $\hat{\eta} \gets (1 - \alpha) \hat{\eta} + \alpha \frac{1}{N} \sum_{n=0}^{N-1} r_n$.
        \STATE Update $\hat{\eta}_- \gets (1 - \alpha) \hat{\eta}_- + \alpha \frac{1}{N} \sum_{n=0}^{N-1} (r_n - \hat{\eta})_-$.
        \STATE Update $\hat{\zeta}_- \gets (1 - \alpha) \hat{\zeta}_- + \alpha \frac{1}{N} \sum_{n=0}^{N-1} (r_n - \hat{\eta})_-^2$.
        \STATE Compute $g(s_n,a_n)$ with Equation~\ref{equ:rew_g} at all timesteps and $\hat{g}$.
        \STATE Compute $\hat{A}_g(s_n,a_n)$ with Equation~\ref{equ:gae} at all timesteps.
        \STATE Update the $\theta$ with Equation~\ref{equ:msvpg}.
        \STATE Update the $\phi$ with  Equation~\ref{equ:val_loss}.
        \ENDFOR
    \end{algorithmic}
\end{algorithm}

\subsection{MSV Trust Region Method}
\label{sec:MSVTRPI}
While PG has a concise form, it often suffers from the difficulty of selecting step-sizes and the sensitivity to initial points in practice, especially when it works with neural networks. To address these drawbacks, trust region method~\shortcite{trpo} is proposed to solve a surrogate problem in a local trust region and perform an approximate policy iteration. 

\subsubsection{Monotonic Improvement Guarantee}
We extend the idea of trust region in the standard MDP into MSV and propose the MSV Trust Region Policy Optimization (MSVTRPO) method. In MSVTRPO, we iteratively solve the problem below

\begin{align} \label{equ:msv_trpo_problem}
    & \max_{\mu_\theta} \mathcal{L}^{\mu}_g(\mu_\theta) \\
    & \ \mathrm{s.t. } \ \mathbb{E}_{s \sim \pi} D_{\rm TV}(\mu_\theta( \cdot \mid s) \parallel \mu( \cdot \mid s) ) \leq \epsilon_\mu, \notag
\end{align}
where
\begin{align*}
    \mathcal{L}^{\mu}_g(\mu_\theta) :=  \mathbb{E}_{s \sim \pi, a \sim {\mu_\theta}} \left[ A_g^\mu(s,a) \right].
\end{align*}

\begin{remark}
The trust region method updates the policy via the direction of maximum derivative (cf. the performance derivative formula in Equation~\ref{equ:derivative formula}), constrained in the proximity policy space with the $TV$-divergence. In contrast, the standard policy iteration scheme updates the policy in the same direction without constraint, which breaks the monotonic improvement for MSV.
\end{remark}

Next, we will show that MSVTRPO enjoys an analogous performance improvement bound. When the trust region is tight enough, i.e., $\epsilon_\mu \to 0$, the lower bound is dominated by the first-order term.

To complete the proof, we need the following lemma to bound the state-action distributions. For a policy $\mu$, we denote the steady state-action distribution as $\rho(s, a) :=\pi(s) \mu(s, a)$. Then we have:
\begin{lemma} \label{lem:rho_diff}
For any two policies $\mu, \mu^\prime \in \Pi$, the difference of their steady state-action distributions $\rho, \rho^\prime$ is bounded by
\begin{align*}
\| \rho^\prime - \rho \|_1 \leq 2 \kappa^\prime \epsilon_\mu.
\end{align*}
\end{lemma}

\begin{proof}

\begin{equation*}
\begin{aligned}
\| \rho^\prime - \rho \|_1 &= \sum_{s, a} | \pi^\prime(s) \mu^\prime( a\mid s) -  \pi(s) \mu( a\mid s) | \\
& \leq  \sum_{s, a} | \pi^\prime(s) \mu^\prime(a \mid s) - \pi(s) \mu^\prime(a \mid s)|  + \sum_{s, a} | \pi(s)  \mu^\prime(a \mid s) - \pi(s) \mu(a \mid s) | \\
&= \sum_s | \pi^\prime(s) - \pi(s) |  + \sum_s \pi(s) \sum_a | \mu^\prime(a \mid s) - \mu(a \mid s) |   \\
&\leq  2 \left( (\kappa^\prime - 1) \epsilon_\mu + \epsilon_\mu \right) = 2 \kappa^\prime \epsilon_\mu,   \\
\end{aligned}
\end{equation*}
where the last inequality follows that $\| \pi^\prime(s) - \pi(s) \|_1 \leq 2(\kappa^\prime -1) \epsilon_\mu$ (see proposition 2 in appendix shown by~\citeR{ijcai2021-385}).
\end{proof}

\begin{theorem}
Let $\mu^\prime$ be the solution to the problem defined by Equation~\ref{equ:msv_trpo_problem}. We have
\begin{equation*}
    \xi^\prime - \xi \geq \mathcal{L}^{\mu}_g(\mu^\prime)  - 2 (\kappa^\prime -1) \epsilon_g \epsilon_\mu - 12 \beta (\kappa^\prime)^2 R_{\rm max}^2 \epsilon_\mu^2,
\end{equation*}
where $\epsilon_g = \max_s | \mathbb{E}_{a \sim \mu^\prime} [A_g^{\mu}(s,a)] |$ and $\kappa^\prime$ is Kemeny's constant under $\mu^\prime$.
\end{theorem}

\begin{proof}
Again, we start our analysis from MSVPDF. Based on Equation~\ref{equ:msv_diff}, we have
\begin{align*}
    \xi_-^\prime - \xi_- &= \mathbb{E}_{s \sim \pi^\prime, a \sim \mu^\prime} [A_f^\mu(s, a)] - \beta \mathbb{E}_{s \sim \pi^\prime, a \sim \mu^\prime} \left[ (r - \eta^\prime)_-^2 - (r - \eta)_-^2  \right] \\
    &= \mathbb{E}_{s \sim \pi^\prime, a \sim \mu^\prime} [A_f^\mu(s, a) + 2\beta \eta_- A_\eta^\mu(s, a)]  - \mathbb{E}_{s \sim \pi^\prime, a \sim \mu^\prime} [2\beta \eta_- A_\eta^\mu(s, a)] \\
    & \quad - \beta \mathbb{E}_{s \sim \pi^\prime, a \sim \mu^\prime} \left[ (r - \eta^\prime)_-^2 - (r - \eta)_-^2  \right] \\
    &= \mathbb{E}_{s \sim \pi^\prime, a \sim \mu^\prime} [A_g^\mu(s, a)] - 2\beta \eta_- (\eta^\prime - \eta) - \beta \mathbb{E}_{s \sim \pi^\prime, a \sim \mu^\prime} \left[ (r - \eta^\prime)_-^2 - (r - \eta)_-^2  \right],
\end{align*}
where the last equation follows the difference formula of average reward, \begin{equation}
    \eta^\prime - \eta = \mathbb{E}_{s \sim \pi^\prime, a \sim \mu^\prime} [A_\eta^\mu(s, a)].
\end{equation} The result indicates that the difference can be separated into two parts: the improvement by optimizing the surrogate problem (the first term), and the discrepancy by the change of $\eta$ (the rest term). The insight of our proof is to show that the first term dominates the difference and the rest terms can be ignored in a tight trust region.

The first term can be tackled with the standard trust region method. 
With the lower bound of average trust region method in Equation~\ref{equ:mean_lower_bound}, we have 
\begin{align} \label{equ:first_term}
    \mathbb{E}_{s \sim \pi^\prime, a \sim \mu^\prime} [A_g^\mu(s, a)] - \mathcal{L}^{\mu}_g(\mu^\prime) \geq  - 2 (\kappa^\prime -1) \epsilon_g \epsilon_\mu.
\end{align}
Now, we need to bound the rest terms. We have
\begin{align*}
    & 2 \eta_- (\eta^\prime - \eta) + \mathbb{E}_{s \sim \pi^\prime, a \sim \mu^\prime} \left[ (r - \eta^\prime)_-^2 - (r - \eta)_-^2  \right]  \\
    &= \mathbb{E}_{s \sim \pi, a \sim \mu} [2(r - \eta)_- (\eta^\prime - \eta)] + \mathbb{E}_{s \sim \pi^\prime, a \sim \mu^\prime} \left[ (r - \eta^\prime)_-^2 - (r - \eta)_-^2  \right] \\
    & = \mathbb{E}_{s \sim \pi^\prime, a \sim \mu^\prime} \left[ (r - \eta^\prime)_-^2 - (r - \eta)_-^2 + 2(r - \eta)_- (\eta^\prime - \eta)\right] \\
    & \quad - 2 (\eta^\prime - \eta) \big( \mathbb{E}_{s \sim \pi^\prime, a \sim \mu^\prime}(r - \eta)_- - \mathbb{E}_{s \sim \pi, a \sim \mu} (r - \eta)_- \big).
\end{align*}
Denote $h:=(r^\prime - \eta^\prime)_{-}^2 - (r^\prime - \eta)_{-}^2 + 2 (r^\prime - \eta)_- (\eta^\prime - \eta)$. Considering all potential cases for the relationship between $\eta, \eta^\prime$ and $h$, we have
\begin{itemize}
    \item If $r \geq \max \{\eta, \eta^\prime \}$, $h = 0$.
    \item If $r < \min \{\eta, \eta^\prime \}$, $h = (r - \eta^\prime)^2 - (r - \eta)^2 + 2 (r - \eta) (\eta^\prime - \eta) = (\eta^\prime - \eta)^2$.
    \item If $\eta \leq r < \eta^\prime$, $h = (r - \eta^\prime)^2 \leq (\eta^\prime - \eta)^2$.
    \item If $\eta^\prime \leq r < \eta$, we denote $c_0 = r - \eta^\prime \geq 0$ and $c_1 = \eta - r >0$. We have  $h = -(r - \eta)^2 + 2(r - \eta) (\eta^\prime - \eta) = c_1^2 + 2 c_0 c_1 \leq (c_0+c_1)^2 = (\eta^\prime - \eta)^2.$
\end{itemize} 
Synthesizing the above results, we conclude $0\leq h \leq (\eta^\prime - \eta)^2$. Thus we have
\begin{equation} \label{equ:rest_terms}
    \mathbb{E}_{s \sim \pi^\prime, a \sim \mu^\prime} \left[ (r - \eta^\prime)_-^2 - (r - \eta)_-^2 + 2(r - \eta)_- (\eta^\prime - \eta)\right] \leq (\eta^\prime - \eta)^2.
\end{equation}

With Lemma~\ref{lem:rho_diff}, we obtain that
\begin{align*}
    |\eta^\prime  - \eta| = |\rho^\prime r - \rho r| \leq \| \rho^\prime - \rho \|_1 R_{\rm max} \leq 2 \kappa^\prime \epsilon_\mu R_{\rm max},
\end{align*}
where the first inequality follows the Hölder's inequality.
Similarly, we have
\begin{align}
    & |\mathbb{E}_{s \sim \pi^\prime, a \sim \mu^\prime}(r - \eta)_- - \mathbb{E}_{s \sim \pi, a \sim \mu} (r - \eta)_-| \\
    &= |\rho^\prime (r - \eta)_- - \rho (r - \eta)_-| \\
    &\leq  \| \rho^\prime - \rho \|_1 R_{\rm max} \label{equ:bound_(r-eta)_} \\
    &\leq  2 \kappa^\prime \epsilon_\mu R_{\rm max}
\end{align}
where Equation~\ref{equ:bound_(r-eta)_} comes from that $0 \leq (r - \eta)_- \leq 2 R_{\rm max}$.
Substituting the previous results into Equation~\ref{equ:rest_terms} and combining with Equation~\ref{equ:first_term}, we arrive at 
\begin{align*}
    \xi_-^\prime - \xi_- &\geq \mathcal{L}^{\mu}_g(\mu^\prime)  - 2 (\kappa -1) \epsilon_g \epsilon_\mu  - \beta |(\eta^\prime -\eta)^2|\\
    & \quad - 2 \beta |\eta^\prime - \eta| \left| \mathbb{E}_{s \sim \pi^\prime, a \sim \mu^\prime}(r - \eta)_- - \mathbb{E}_{s \sim \pi, a \sim \mu} (r - \eta)_- \right| \\
    & \geq \mathcal{L}^{\mu}_g(\mu^\prime) - 2 (\kappa^\prime -1) \epsilon_g \epsilon_\mu - 12 \beta (\kappa^\prime)^2 R_{\rm max}^2 \epsilon_\mu^2.     
\end{align*}


\end{proof}

\subsubsection{Implementation details}
At the end of this subsection, we address some implementation issues of MSVTRPO. First of all, in practice, we replace the ${\rm TV}$-divergence with ${\rm KL}$-divergence as most trust region methods do. Since $D_{\rm TV}(p \parallel q) \leq \sqrt{D_{\rm KL}(p \parallel q)/2}$, the theoretical results are still applicable for the practical algorithms.

In the tabular case, where the state and action spaces are finite and discrete, it is enough to parameterize the policy tabularly. The previous analysis of TRPO~\shortcite{abdolmaleki2018maximum} shows that Equation~\ref{equ:msv_trpo_problem} enjoys a closed form solution:
\begin{equation*}
    \mu^\prime(\cdot \mid s, a) \propto \mu(\cdot \mid s, a) \exp \left( \dfrac{A_g^\mu(s,a)}{\upsilon^*} \right),
\end{equation*}
where $\upsilon^*$ can be obtained by solving the dual problem
\begin{equation*}
\min_\upsilon \mathcal{L}(\upsilon) := \upsilon \epsilon_\mu + \upsilon \sum_s \pi(s) \log \sum_a \mu\left(a \mid s\right) \exp \left(\frac{A_g^\mu(s, a)}{\upsilon}\right).
\end{equation*}
With a known MDP, we name this iterative procedure MSV Trust Region Policy Iteration (MSVTRPI). As aforementioned in Section~\ref{sec:PA}, PI is not available for MSV. Nevertheless, we can do MSVTRPI as an alternative. When $\epsilon_\mu \to \infty$, it degrades to the standard PI without the monotonic improvement guarantee. 

\begin{algorithm}[tb] 
\caption{MSVPO}  
\label{alg:MSVPPO}  
    \textbf{Input}: $\alpha, \beta, K, N, M$
    \begin{algorithmic}[1]
        \STATE Initialize the policy with $\theta$ and the value with $\phi$ randomly.
        \STATE Set $\hat{\eta} = 0$, $\hat{\eta}_- = 0$, $\hat{\zeta}_- = 0$.
        \FOR {$k=1,2,\cdots, K$} 
        \STATE Execute policy $\mu_{\theta}$ for $N$ times to collect $\{(s_n, a_n, r_n, s_{n+1})\}_{n=0}^{N-1}$.
        \STATE $\hat{\eta} \gets (1 - \alpha) \hat{\eta} + \alpha \frac{1}{N} \sum_{n=0}^{N-1} r_n$.
        \STATE $\hat{\eta}_- \gets (1 - \alpha) \hat{\eta}_- + \alpha \frac{1}{N} \sum_{n=0}^{N-1} (r_n - \hat{\eta})_-$.
        \STATE $\hat{\zeta}_- \gets (1 - \alpha) \hat{\zeta}_- + \alpha \frac{1}{N} \sum_{n=0}^{N-1} (r_n - \hat{\eta})_-^2$.
        \STATE Compute $g(s_n,a_n)$ with Equation~\ref{equ:rew_g} at all timesteps and $\hat{g}$.
        \STATE Compute $\hat{A}_g(s_n,a_n)$ with Equation~\ref{equ:gae} at all timesteps.
        \STATE Update the $\theta$ with equation \ref{equ:J_pi} for $M$ epochs.
        \STATE Update the $\phi$ with  Equation~\ref{equ:val_loss} for $M$ epochs.
        \ENDFOR
    \end{algorithmic}
\end{algorithm}

In the model-free case with large state and action spaces, we recommend solving the surrogate loss proposed by PPO~\shortcite{ppo}, for its stable performance and fast computing with neural networks. Formally, instead of optimizing the problem in Equation~\ref{equ:msv_trpo_problem}, we maximizing the clipping objective
\begin{equation}\label{equ:J_pi}
    \mathcal{L}_\mu^{\rm CLIP}(\theta) :=
    \frac{1}{N }\sum_{n=0}^{N-1} \left[
    \min \left(\omega_n(\theta) \hat{A}_g(s_n, a_n), 
    \text{clip}( \omega_n(\theta),
    1-\varepsilon,
    1+\varepsilon)\hat{A}_g(s_n, a_n) \right) \right],
\end{equation}
where $\omega_n(\theta) = \frac{\mu_\theta(a_n \mid s_n)}{\mu(a_n \mid s_n)}$ is the importance sampling ratio. Since we consider the long-run average performance in this paper, GAE is not applicable directly. Thus, we adopt the average value constraint (AVC) proposed by~\shortciteA{ijcai2021-385} to stabilize the value learning. The full algorithm, named by MSV Policy Optimization (MSVPO) is presented in Algorithm~\ref{alg:MSVPPO}.

\section{Experiments}

In the previous sections, we analyze the properties of MSV problem and find that it can be solved by iteratively optimizing a surrogate reward function $g$ (cf. Equation~\ref{equ:rew_g}). We also propose two methods to solve the MSV problem in the parameterized policy space. 

To validate the effectiveness of our proposed methods in solving MSV problem, we conduct a series of experiments to answer the corresponding questions:
\begin{itemize}
    \item Is the MSV really optimized by the surrogate reward function $g$? Specifically, what is the difference from optimizing $g$ instead of $f$?
    \item What is the difference between the MV~\shortcite{xia2020risk} and MSV criteria?
    \item Does the proposed algorithms work well with the current deep RL algorithms?
\end{itemize}

\subsection{Bandit Problem}

\begin{figure*}[htbp]
  \centering
    \begin{subfigure}{0.45\textwidth}
        \centering
        \includegraphics[width=\linewidth]{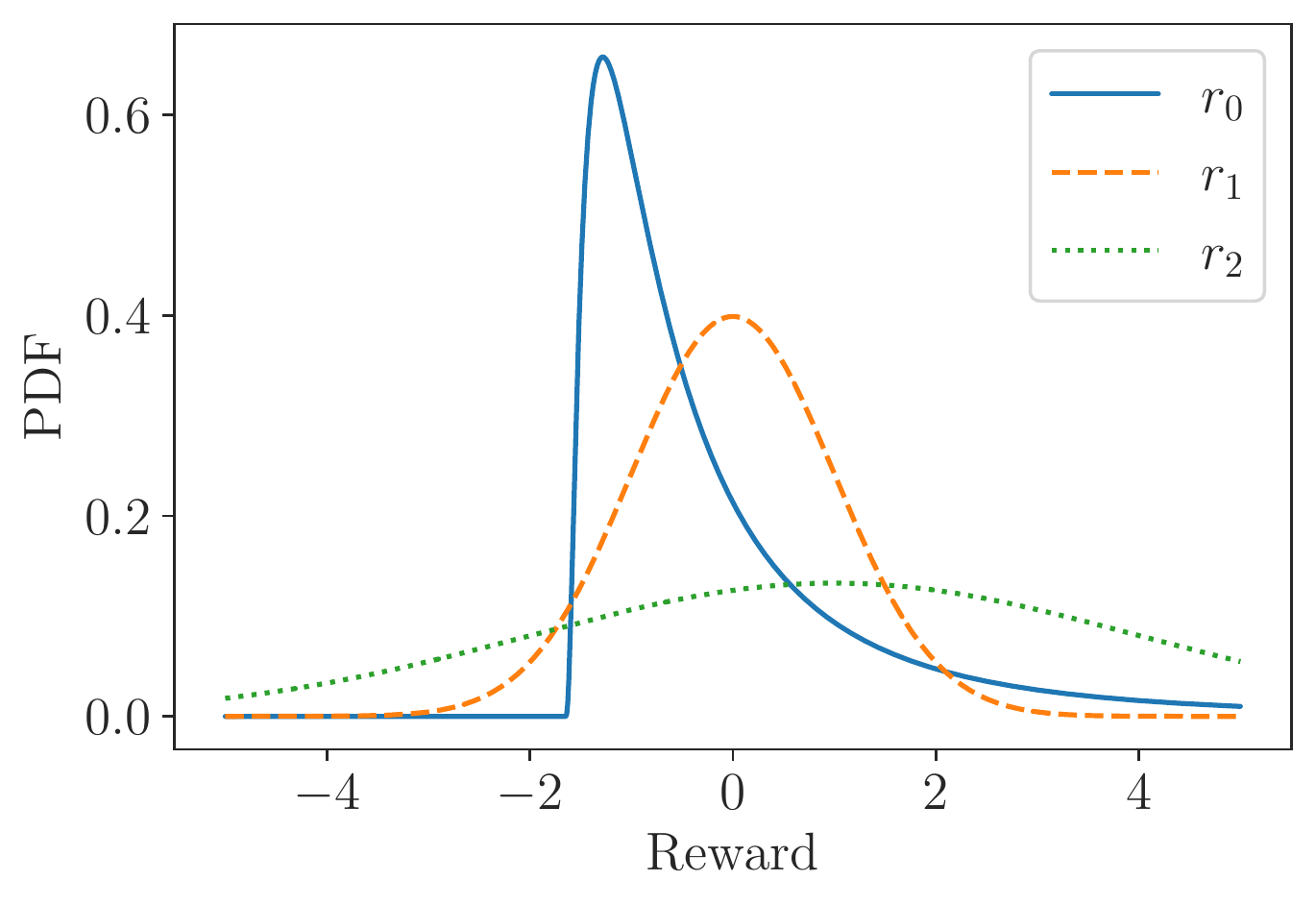}
        \caption{Reward distributions.}
    \end{subfigure}   
    \begin{subfigure}{0.45\textwidth}
        \centering
        \includegraphics[width=\linewidth]{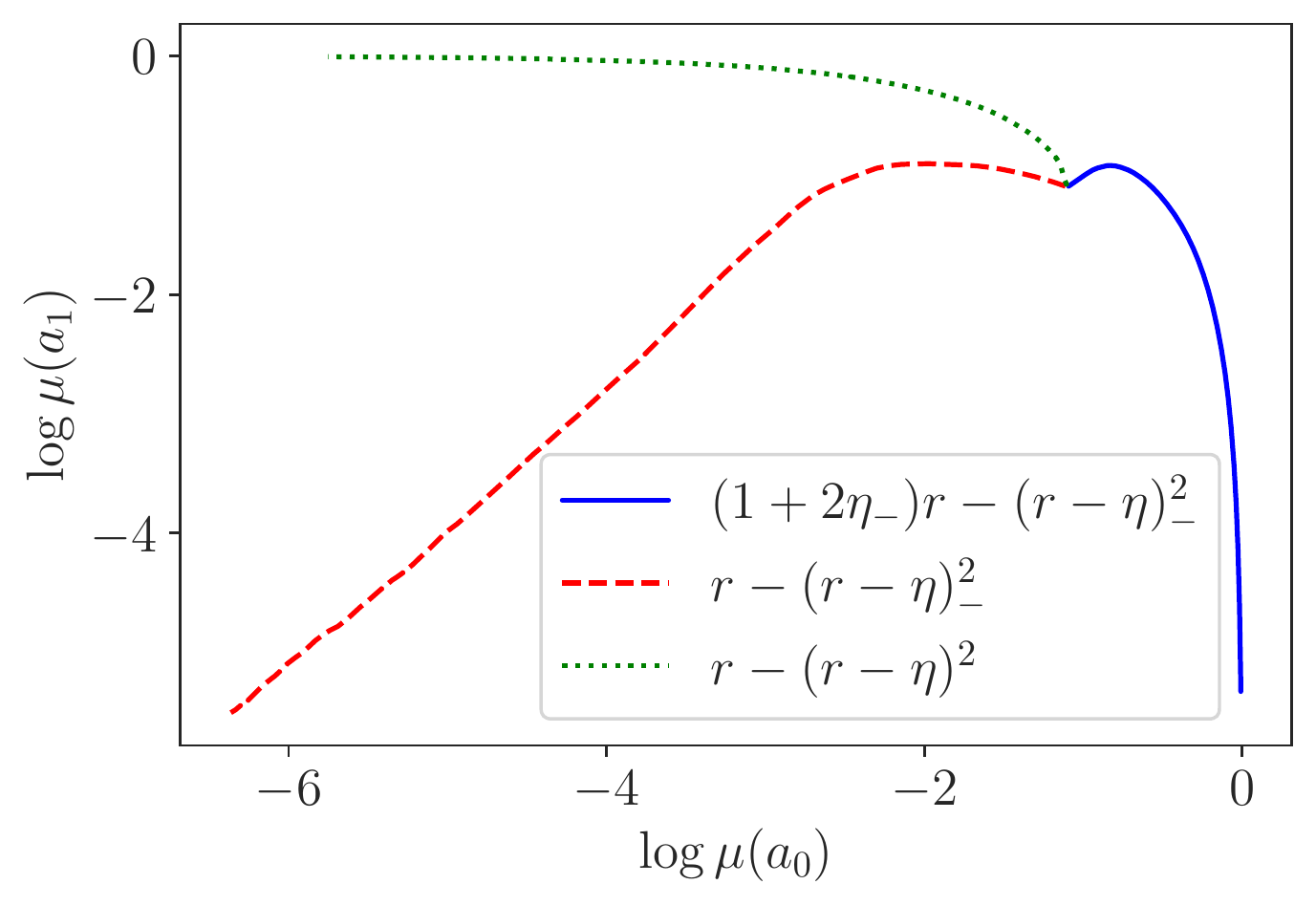}
        \caption{Policies paths.}
    \end{subfigure}
    \caption{The bandit problem. (a) Reward distributions in the bandit problem. (b) Policies paths in the bandit problem. The paths are shown in the logarithmic parameter space.}
     \label{fig:bandit}
\end{figure*}



We start with a simple bandit problem. In this problem, there are three actions with only a single state. Different actions result in different rewards following the distributions shown in Figure~\ref{fig:bandit}(a). Specifically, we have $r_0$ sampled from a shifted $\text{LogNormal}(0, 1)$ distribution, of which the mean is shifted to zero. If we choice $a_1$, we will obtain $r_1 \sim N(0, 2^2)$. Otherwise, we will have $r_2 \sim N(1, 3^2)$. Obviously, we have three different risk preference actions. When we fix $\beta=1$ in MV and MSV, the agent should always choose $a_0$ if it optimizes the MSV criterion, and choose $a_1$ if it optimizes the MV criterion. The $a_2$ has the highest outcome, which is preferred by risk-neutral agents.

We compare three different agents, which optimize different reward functions. The first one optimizes $g = (1 + 2 \eta_-) r - (r- \eta)_-^2$ (cf. Equation~\ref{equ:rew_g}), which is the derived reward function with $\beta=1$ in this work. The second one optimizes $f = r - (r- \eta)_-^2$ (cf. Equation~\ref{equ:f}), which is the Monte-Carlo return of MSV. We further consider a third agent which optimizes $r - (r - \eta)^2$~\shortcite{xia2020risk}, an MV objective to illustrate the difference between MSV and MV problems. All the agents use policy gradient with a parameterized policy initialized as a uniform one.

To visualize the learning process, we plot the curves in the logarithmic parameter space, as shown in Figure~\ref{fig:bandit}(b). Since $\sum_i \mu(a_i) = 1$, $\mu(a_2)$ is ignored in the figure. As expected, the learning curve of the first agent (blue solid curve) approaches $(0, -\infty)$, meaning that it always chooses $a_0$ finally. Similarly, the third agent (green dotted curve) also chooses $a_1$ correspondingly. Interestingly, the second agent (red dashed curve), which optimizes the Monte-Carlo return of MSV, finally converges to choose $a_2$. The result tells us optimizing the reward $f = r - \beta(r-\eta)_-^2$ cannot optimize the MSV objective even in such a simple problem. This reflects the most essential difference between the optimization of policy-dependent reward and other problems. As discussed in Section~\ref{sec:PA}, to optimize a problem with a policy-dependent reward function, we must consider the perturbation of the mean, at least in MSV problems.

\subsection{Portfolio Management}

\begin{figure}[tbp]
    \centering
    \includegraphics[width=0.8\linewidth]{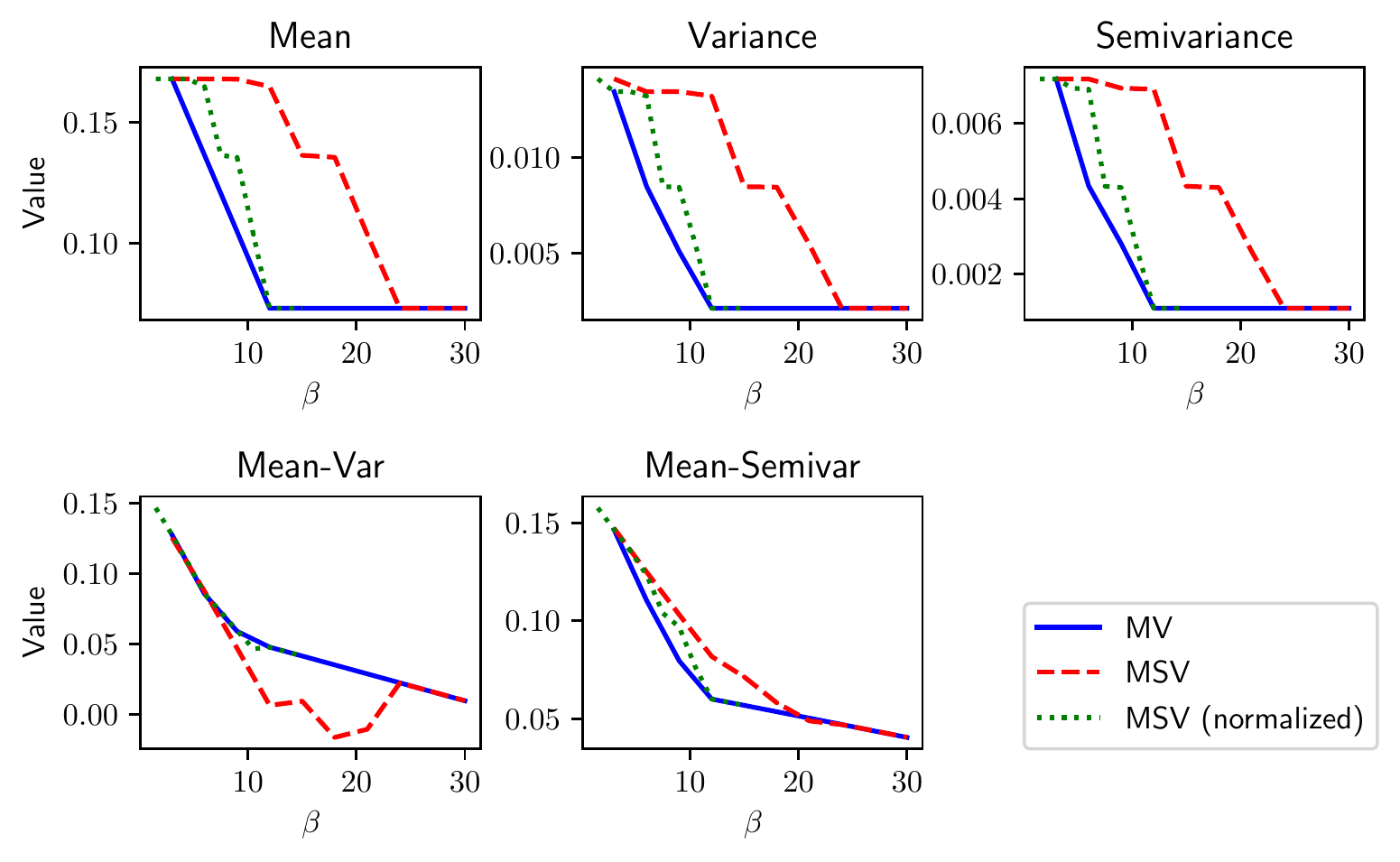}
    \caption{Comparison of MSVTRPI and MVPI in the portfolio management problem. The normalized MSV means $\beta$ is doubled in comparison.}
    \label{fig:pm_curves_full}
\end{figure}

In this part, we compare the performances of MSV- and MV-optimal policies in a portfolio management problem. We need to manage two independent assets and cash. At the stage $t$, the gain of the $i$-th asset is denoted by $x_{i, t} \in \{-0.2, -0.1, \dots, 0.5\}$, which transits according to a transition probability matrix (described in Appendix~\ref{app:pm_setup}). The action space is defined as $\mathcal{A} = \{ (w_{1, t}, w_{2, t}) \mid \sum_{i=1,2} w_{i,t} \leq 1, w_{i, t} \in \{0, 0.2, \dots, 1\} \}$, where $w_{i, t}$ is the weight of current portfolio on the $i$-th asset. Let $w_{0, t} = 1 - w_{1, t} -  w_{2, t}$ denote the partition of cash in current portfolio and $x_0$ denote the return of cash. The reward function is defined as $r_t = w_{0, t} x_0 + \sum_{i=1,2} w_{i, t} x_{i, t}  - \sum_{i=1,2} |w_{i, t} - w_{i, t-1}| c$, where $c$ is the transition cost. The state is defined as $s_t = (x_{1, t}, x_{2, t}, w_{0, t}, w_{1, t})$. Hence, $|\mathcal{A}| = 21$ and $|\mathcal{S}|=1344$. 

For the MSV, we optimize the policy with the MSV trust region policy iteration (MSVTRPI) (see Section~\ref{sec:MSVTRPI} for details), which aims to maximize $\xi_-^\mu = \eta^\mu - \beta \zeta_-^\mu$. We parameterize the policy in the softmax form as $\mu_\theta( a \mid s) := {\rm softmax}(\theta(s, a))$, where $\theta \in \mathbb{R}^{|S||A|} $ are the ``logic values''. For MV, we optimize the policy with the mean-variance policy iteration (MVPI) proposed by~\shortciteA{xia2020risk}, which maximizes $\xi^\mu = \eta^\mu - \beta \zeta^\mu$.

\begin{figure}[tbp]
    \centering
    \includegraphics[width=0.6\linewidth]{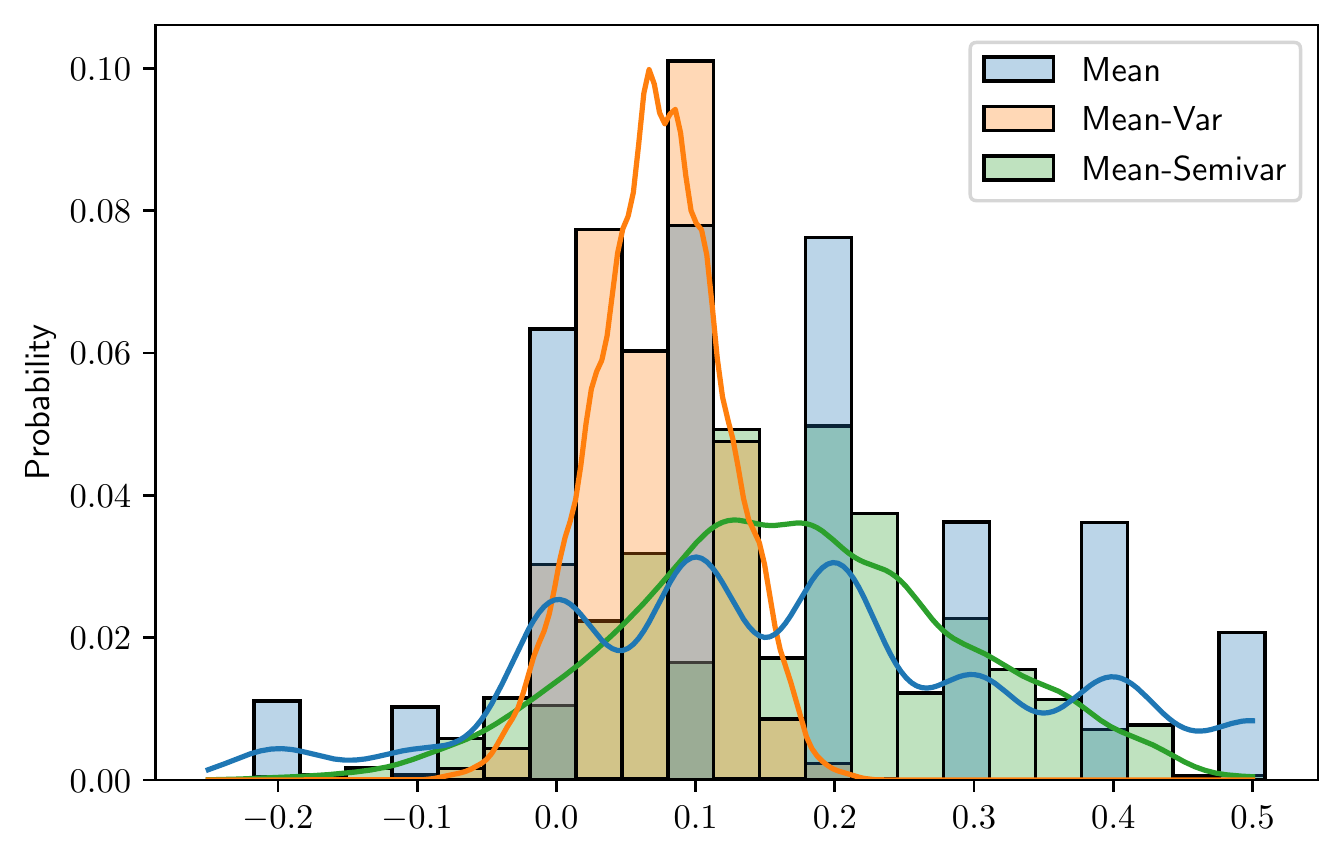}
    \caption{Reward distribution in the portfolio management problem. The policy optimizing MSV achieves $\eta=0.168,\zeta=0.014,\zeta_-=0.006$. As a comparison, the policy optimizing MV achieves $\eta=0.073,\zeta=0.002,\zeta_-=0.001$.}
    \label{fig:pm_dist}
\end{figure}

We change the risk preference parameter $\beta$ and compare the MSVTRPI and MVPI. We depict the result in Figure~\ref{fig:pm_curves_full}, showing that with a fixed $\beta$, optimizing MSV always results in a larger return than that of MV. Besides, MV is more sensitive than MSV in terms of $\beta$, meaning that a small change of $\beta$ will lead to a quick drop in both the return and risk. To better compare MSV and MV, we also show the ``normalized'' results of MSV, where we double $\beta$ to provide the same penalty strength as MV. The result shows the normalized MSV also outperforms MV in terms of the average reward, illustrating that MSV is more plausible than MV. We demonstrate the reward distributions in Figure~\ref{fig:pm_dist} with $\beta=10$. It shows that MSV maintains high returns while avoiding large losses. In contrast, optimizing MV may be too conservative, as the upside rewards cause more volatility in this problem.

\subsection{Robotic Control}

To demonstrate the effectiveness of our proposed method in more general problem setups, we implement a ``deep'' variant algorithm named mean-semivariance policy optimization (MSVPO), which is based on the recently developed method APO~\shortcite{ijcai2021-385} for average-reward RL problems.

We evaluate MSVPO in the continuous control benchmark MuJoCo~\shortcite{mujoco} with OpenAI gym~\shortcite{brockman2016openai} as the interface. Since the original setup of MuJoCo is not suitable for the long-run average setting, we slightly modify the experimental protocol. In most of MuJoCo tasks, the agent will be terminated if it reaches any unsafe state, such as falling down. In that case, we will reset the system and add an extra cost to the terminal state. Different from other works focusing on the average episode returns, we are interested in the long-run average and semivariance of the steady reward distribution. To further increase the risk in the test scenarios, we add some noise to the agent outputs, i.e., the real action taken by the environment is $a_t + \epsilon$, where $\epsilon \sim N(0, \sigma^2)$. We call $\sigma$ as the noise level of the modified MuJoCo tasks.

\begin{figure}[tbp]
    \centering
    \includegraphics[width=\linewidth]{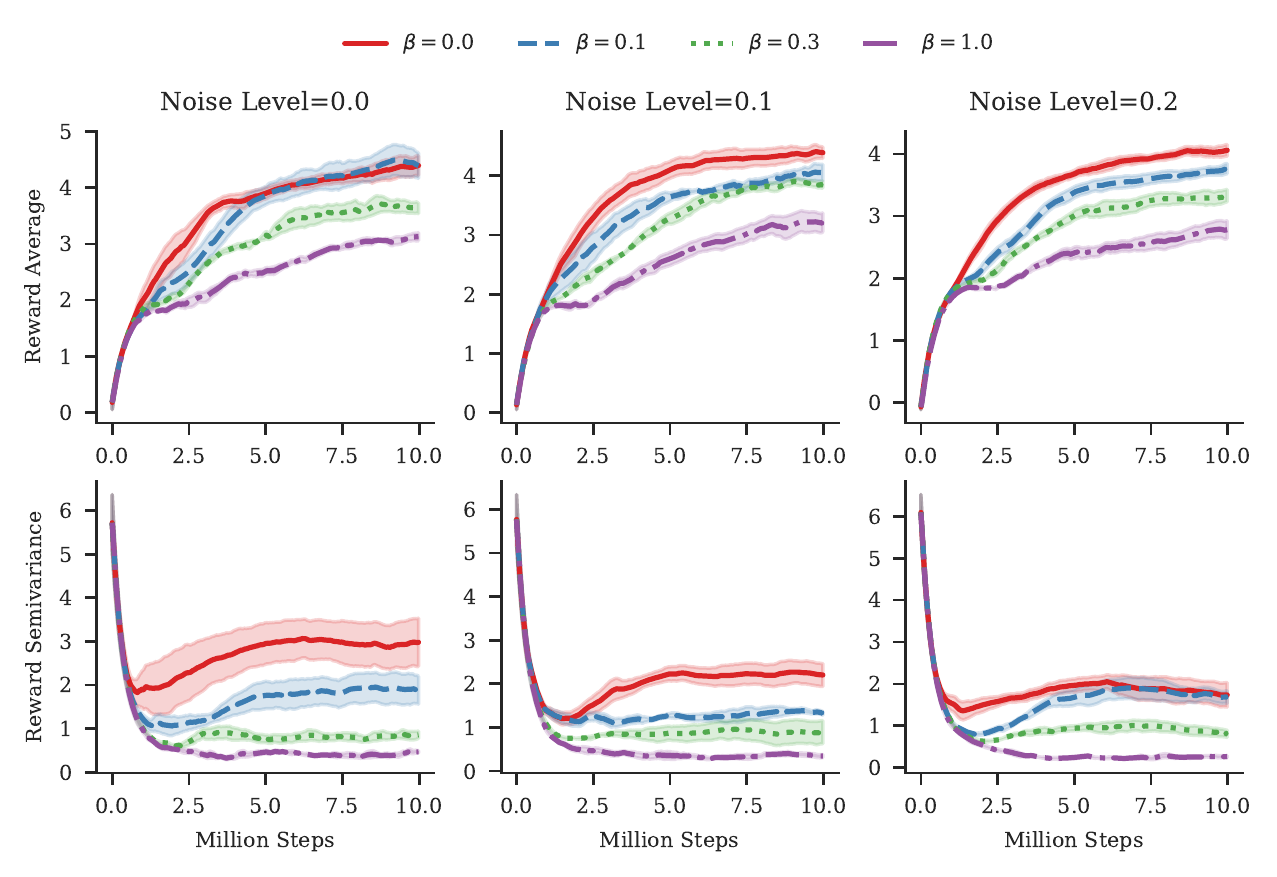}
    \caption{Training curves of Walker2d with noise. Each curve is averaged over 10 random seeds and shaded by the standard deviation.}
    \label{fig:traning_curves}
\end{figure}

We evaluate MSVPO with different $\beta$'s in the noisy Walker2d with different noise levels. When the agent falls, we penalize it with an extra cost -10 and reset the system. As shown in Figure~\ref{fig:traning_curves}, the choice of different $\beta$'s achieves the trade-off between the average and semivariance. In the noiseless environment (noise level = 0), we interestingly find that risk-averse policy ($\beta=0.1$) achieves competitive average reward with lower semivariance. It indicates that in complex scenes, optimizing a risk-averse metric may generate more robust policies with better performances compared with a risk-neutral one.

\begin{figure}[htbp]
    \centering
    \includegraphics[width=0.7\linewidth]{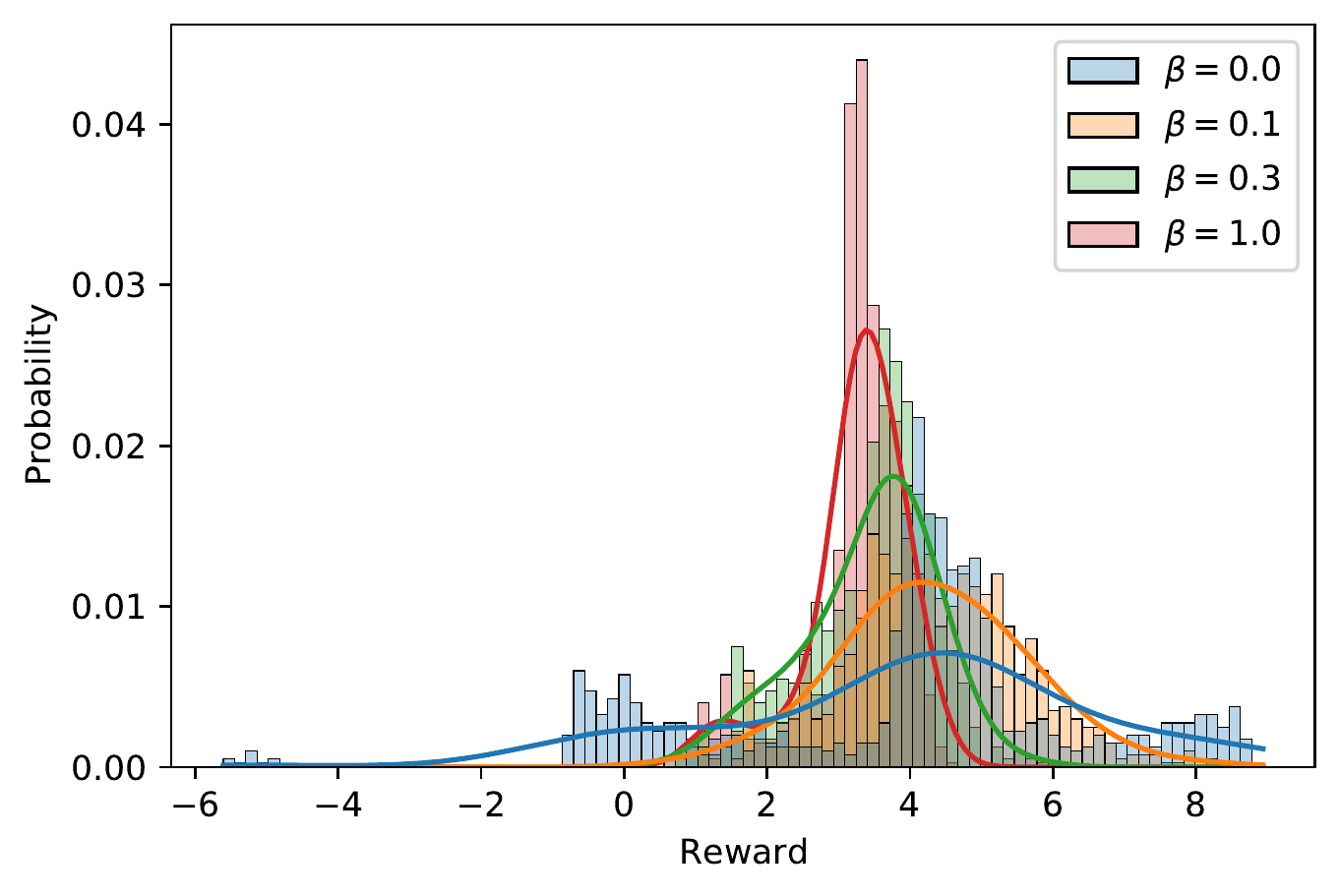}
    \caption{Reward distribution of Walker2d with noise.}
    \label{fig:walker2d_dist}
\end{figure}

To better understand the performance difference with different risk preference policies, we visualize the reward distributions of typical agents in Figure~\ref{fig:walker2d_dist}, where each agent of noise level 0.1 is evaluated for 1000 steps. We can see that risk-averse policies successfully avoid unsafe states.
Meanwhile, the agent uses smaller steps forward with the risk parameter $\beta$ increasing. 
Instead, the risk-neutral agent tends to take the risk of falling for larger gains. 

\section{Conclusion}
This paper discusses how to optimize the mean-semivariance criterion for the steady reward of MDPs and RL, which is an alternative risk measure of mean-variance. The semivariance is a more reasonable measure than the variance in general scenarios, as it only penalizes the downside risk. We utilize PA theory to derive the performance difference formula and optimize MSV with data-driven approaches. We develop two algorithms for MSV based on PA theory, following the policy gradient theory and the trust region theory, respectively. We also demonstrate the effectiveness of the proposed algorithms in different problems, showing the risk-averse performance of MSV policy. We point out that the application of the proposed two-stage optimization framework for risk measures is not limited to MSV. We hope our work can promote the applications of data-driven approaches in risk-sensitive environments of MDPs and RL.

\acks{This work is funded by the National Natural Science Foundation of China (No. U1813216, 62192751, 61425027, 62073346, 11931018, U1811462), the National Key Research and Development Project of China under Grant 2017YFC0704100 and Grant 2016YFB0901900, in part by the 111 International Collaboration Program of China under Grant BP2018006, BNRist Program (BNR2019TD01009), the National Innovation Center of High Speed Train R\&D project (CX/KJ-2020-0006), the Guangdong Province Key Laboratory of Computational Science at the Sun Yat-Sen University (2020B1212060032) and the Guangdong Basic and Applied Basic Research Foundation (2021A1515011984).
}

\newpage
\appendix
\section{Brief Review of Perturbation Analysis theory} \label{app:PA}
Consider an ergodic MDP with transition matrix $P$ (induced by some policy $\mu$), where $P(s^\prime \mid s)$ is the transition probability from $s$ to $s^\prime$. We also consider a corresponding reward function $r$, where $r(s)$ is the reward expectation at $s$. We are interested in the average performance $\eta = \pi r$, where $\pi$ denotes the steady-state distribution. The Perturbation Analysis (PA) theory~\cite{cxr} captures how the performance changes if the policy (or system parameters $P$ and $r$) has perturbations.

\begin{theorem}[Performance difference formula]
 For two ergodic MDPs with $P$ and $P^\prime$, we have
 \begin{equation*}
     \eta^\prime - \eta = \pi [(P^\prime-P)V + r^\prime - r],
\end{equation*}
where $V$ is the value function (called potential function in PA) for the system with $P$.
\end{theorem}
The value function satisfies the Poisson equation $(I - P)g + \eta e= r$, where $I$ denotes the identity matrix and $e$ is the unit vector.

\begin{theorem}[Performance derivative formula]
 Consider another MDP with $P^\nu = P + \Delta P=(1-\nu)P + \nu P^\prime$ and $r^\nu=r+ \nu \Delta r=(1-\nu)r + \nu r^\prime$. We have
 \begin{equation*}
     \left. \frac{{\rm d} \eta}{{\rm d} \nu} \right|_{\nu=0} = \pi [(\Delta P) V + \Delta r].
\end{equation*}
\end{theorem}

\section{Alternative Proof of MSVPG}
This proof follows the similar derivation of~\shortciteA[Chapter 13]{sutton2018reinforcement}.
We first derive the policy gradient of $\zeta_-$, and give the complete form of MSV gradient by $\nabla_\theta \xi_- = \nabla_\theta \eta - \beta \nabla_\theta \zeta_-$. Taking the gradient of $V_{\zeta_-}^{\mu}$ for any arbitrary $s \in \mathcal{S}$, we have
\begin{align*}
&\nabla_\theta V_{\zeta_-}^{\mu}(s) \\
&= \nabla_\theta \Big[\sum_{a} \mu(a \mid s) Q_{\zeta_-}^{\mu}(s, a)\Big]\\
&=\sum_{a}\Big[ \nabla_\theta \mu(a \mid s) Q_{\zeta_-}^{\mu}(s, a)+\mu(a \mid s) \nabla_\theta Q_{\zeta_-}^{\mu}(s, a)\Big] \\
&=\sum_{a}\Big[\nabla_\theta \mu(a \mid s) Q_{\zeta_-}^{\mu}(s, a) +\mu(a \mid s) \nabla_\theta \sum_{s^{\prime}} P\left(s^{\prime} \mid s, a\right) \big( (r-\eta)^2_- - \zeta_-+V_{\zeta_-}^{\mu}\left(s^{\prime}\big)\right) \Big] \\
&=\sum_{a}\Big[ \nabla_\theta \mu(a \mid s) Q_{\zeta_-}^{\mu}(s, a)+ \mu(a \mid s) \sum_{s^{\prime}} P(s^{\prime} \mid s, a)\big( -2(r-\eta)_- \nabla_\theta\eta - \nabla_\theta \zeta_-+ \nabla_\theta V_{\zeta_-}^{\mu}\left(s^{\prime}\big)\right) \Big].
\end{align*}
Rephrasing the equation above, we obtain
\begin{align*}
&\nabla_\theta \zeta_- =\\
&\sum_{a}\Big[\nabla_\theta \mu(a \mid s) Q_{\zeta_-}^{\mu}(s, a) + \mu(a \mid s)\sum_{s^{\prime}} P\left(s^{\prime} \mid s, a\right) \left(\nabla_\theta V_{\zeta_-}^{\mu}\left(s^{\prime}\right) -2(r-\eta)_- \nabla_\theta\eta  \right) \Big] - \nabla_\theta V_{\zeta_-}^{\mu}(s).      
\end{align*}
Taking the expectation under $\pi$ for both sides, we have
\begin{align}
&\nabla_\theta \zeta_- \notag \\
&=  \sum_{s} \pi(s) \sum_{a}\Big[\nabla_\theta \mu(a \mid s) Q_{\zeta_-}^{\mu}(s, a) + \notag  \mu(a \mid s) \sum_{s^{\prime}} P\left(s^{\prime} \mid s, a\right) \big( \nabla_\theta V_{\zeta_-}^{\mu}\left(s^{\prime}\big) -2 (r-\eta)_- \nabla_\theta\eta \right) \Big] \notag \\
& \quad -\sum_{s} \pi(s) \nabla_\theta V_{\zeta_-}^{\mu}(s) \notag  \\
&= \sum_{s} \pi(s) \sum_{a} \nabla_\theta \mu(a \mid s) Q_{\zeta_-}^{\mu}(s, a) \notag +\sum_{s^{\prime}} \sum_{s} \pi(s) \sum_{a} \mu(a \mid s) P\left(s^{\prime} \mid s, a\right) \nabla_\theta V_{\zeta_-}^{\mu}\left(s^{\prime}\right) \notag \\
& \quad - \sum_{s} \pi(s) \sum_{a} \mu(a \mid s) \sum_{s^{\prime}} 2(r-\eta)_- \nabla_\theta\eta  -\sum_{s} \pi(s) \nabla_\theta V_{\zeta_-}^{\mu}(s). \label{equ:111}
\end{align}    

By the definitions of $\pi$ and $\eta_-$, we have
\begin{align*}
    \pi\left(s^{\prime} \right) &= \sum_{s} \pi(s) \sum_{a} \mu(a \mid s) P\left(s^{\prime} \mid s, a\right),  \\
    \eta_- &= \sum_{s} \pi(s) \sum_{a} \mu(a \mid s) \sum_{s^{\prime}} (r-\eta)_-.
\end{align*}
Substituting into the Equation~\ref{equ:111}, we have 
\begin{align*}
\nabla_\theta \zeta_- &= \sum_{s} \pi(s) \sum_{a} \nabla_\theta \mu(a \mid s) Q_{\zeta_-}^{\mu}(s, a) + \sum_{s^{\prime}} \pi\left(s^{\prime}\right) \nabla_\theta  V_{\zeta_-}^{\mu}\left(s^{\prime}\right) - 2 \eta_- \nabla_\theta\eta -\sum_{s} \pi(s) \nabla_\theta V_{\zeta_-}^{\mu}(s) \\
&= \sum_{s} \pi(s) \sum_{a} \nabla_\theta \mu(a \mid s) Q_{\zeta_-}^{\mu}(s, a) - 2 \eta_- \nabla_\theta\eta \\
&= \sum_{s} \pi(s) \sum_{a} \nabla_\theta \mu(a \mid s) Q_{\zeta_-}^{\mu}(s, a) - 2 \eta_- \sum_{s} \pi(s) \sum_{a} \nabla_\theta \mu(a \mid s) Q_\eta^\mu(s, a) \\
&= \sum_{s} \pi(s) \sum_{a} \nabla_\theta \mu(a \mid s) \Big[Q_{\zeta_-}^{\mu}(s, a) - 2 \eta_-  Q_\eta^\mu(s, a)\Big].
\end{align*}
Finally, applying the trick $\nabla \log \mu = \nabla \mu / \mu$, we have 
\begin{equation*}
    \nabla_\theta \zeta_- = \mathbb{E}_{s \sim \pi, a \sim \mu} \Big[Q_{\zeta_-}^{\mu}(s, a) - 2 \eta_-  Q_\eta^\mu(s, a)\Big].
\end{equation*}
Thus, the MSVPG is given by
\begin{equation*}
    \nabla_\theta \xi- = \mathbb{E}_{s \sim \pi, a \sim \mu} \Big[(1 + 2\eta_-)  Q_\eta^\mu(s, a) - \beta Q_{\zeta_-}^{\mu}(s, a)\Big].
\end{equation*}

\section{Experiment Details}
\subsection{The Setup of Portfolio Management Problem} \label{app:pm_setup}
    The return of cash $x_0 = 0.01$.
    The transition cost $c = 0.05$.

\begin{table}[htbp]
	\centering
	\caption{The transition matrix of asset 1}
	\begin{tabular}{rrrrrrrrr}
		\toprule  
		$x_1$ & -0.2 & -0.1 & 0 & 0.1 & 0.2 & 0.3 & 0.4 & 0.5 \\ 
		\cmidrule(r){2-9}
        -0.2 & 0.09 & 0.05 & 0.25 & 0.24 & 0.18 & 0.05 & 0.10 & 0.04 \\
        -0.1 & 0.05 & 0.02 & 0.33 & 0.22 & 0.17 & 0.09 & 0.06 & 0.06 \\
        0    & 0.04 & 0.03 & 0.26 & 0.24 & 0.18 & 0.07 & 0.12 & 0.06 \\
        0.1 & 0.04 & 0.04 & 0.20 & 0.28 & 0.26 & 0.08 & 0.03 & 0.07 \\
        0.2 & 0.00 & 0.02 & 0.16 & 0.24 & 0.27 & 0.11 & 0.15 & 0.05 \\
        0.3 & 0.07 & 0.02 & 0.16 & 0.19 & 0.25 & 0.14 & 0.12 & 0.05 \\
        0.4 & 0.02 & 0.04 & 0.14 & 0.19 & 0.18 & 0.20 & 0.17 & 0.06 \\
        0.5 & 0.03 & 0.03 & 0.09 & 0.19 & 0.23 & 0.15 & 0.14 & 0.14 \\
		\bottomrule  
	\end{tabular}
\end{table}

\begin{table}[htbp]
	\centering
	\caption{The transition matrix of asset 2}
	\begin{tabular}{rrrrrrrrr}
		\toprule  
		$x_2$ & -0.2 & -0.1 & 0 & 0.1 & 0.2 & 0.3 & 0.4 & 0.5 \\ 
		\cmidrule(r){2-9}
        -0.2 & 0.13 & 0.10 & 0.08 & 0.09 & 0.20 & 0.36 & 0.02 & 0.02 \\
        -0.1 & 0.06 & 0.11 & 0.09 & 0.12 & 0.17 & 0.37 & 0.04 & 0.04 \\
        0    & 0.01 & 0.06 & 0.12 & 0.15 & 0.25 & 0.35 & 0.02 & 0.04 \\
        0.1 & 0.06 & 0.06 & 0.12 & 0.15 & 0.22 & 0.34 & 0.01 & 0.04 \\
        0.2 & 0.02 & 0.04 & 0.09 & 0.24 & 0.23 & 0.32 & 0.04 & 0.02 \\
        0.3 & 0.04 & 0.07 & 0.11 & 0.20 & 0.26 & 0.27 & 0.03 & 0.02 \\
        0.4 & 0.10 & 0.11 & 0.13 & 0.16 & 0.17 & 0.20 & 0.04 & 0.09 \\
        0.5 & 0.01 & 0.10 & 0.30 & 0.21 & 0.16 & 0.16 & 0.00 & 0.06 \\
		\bottomrule  
	\end{tabular}
\end{table}

\subsection{Hyper-parameters of MSVPO}
\label{sec:hyperparameter}

\begin{table}[H]
\centering
\begin{tabular}{lr}
\toprule
Hyper-parameter & Value \\
\midrule
 Network learning rate $\beta$ & 3e-4 \\
 Network hidden sizes  & [64, 64] \\
 Activation function & Tanh \\
 Optimizer & Adam \\
 Batch size & 256 \\
 Gradient Clipping & 10 \\
 Clipping parameter $\varepsilon$ & 0.2 \\
 Optimization Epochs $M$ & 10 \\
 GAE parameter $\lambda$ & 0.95 \\
 Average Value Constraint Coefficient in APO~\shortcite{ijcai2021-385} $\nu$ & 0.3 \\
\bottomrule
\end{tabular}
\caption{Hyper-parameters sheet}
\label{tab:hyper}
\end{table}


\bibliographystyle{theapa}
\bibliography{sample}

\end{document}